\documentclass[10pt]{article} % For LaTeX2e

\usepackage[accepted]{rlj} % Should be uncommented for the camera-ready
\usepackage{amssymb}            % Defines common symbols like \mathbb R
\usepackage{mathtools}          % Extends amsmath, providing common math tools
\usepackage{mathrsfs}           % Enables \mathscr, which can work in cases that \mathcal does not
\usepackage{graphicx}           % For including images
\usepackage{subcaption}         % Allows for the use of subfigures and subcaptions
\usepackage[space]{grffile}     % For spaces in image names
\usepackage{url}                % For displaying URLs
\usepackage{lipsum}  

\usepackage{amsmath}
\usepackage{amsthm}
\newtheorem{theorem}{Theorem}

\newtheorem{remark}[theorem]{Remark}
\usepackage{algorithm}
\usepackage{algpseudocode}
\usepackage{graphicx}
\usepackage{subcaption}
\usepackage{multirow}
\usepackage{booktabs}
\usepackage{capt-of}
\usepackage{wrapfig}

\newcommand{\E}{\mathbb{E}}

\title{GCT-MARL: Graph-Based Contrastive Transfer for Sample-Efficient Cooperative Multi-Agent Reinforcement Learning}

\setrunningtitle{GCT-MARL}

\author{Animesh Animesh\textsuperscript{1,$\dagger$}, Satheesh K Perepu\textsuperscript{2}, Kaushik Dey\textsuperscript{2}}

\emails{animesh.sachan24794@kgpian.iitkgp.ac.in, \\ \{perepu.satheesh.kumar,deykaushik\}@ericsson.com}

\affiliations{
$^{1}$\textbf{Department of Artificial Intelligence, Indian Institute of Technology Kharagpur, India}\\
$^{2}$\textbf{Ericsson Research, Bangalore, India}
\par % If including additional comments like below, use \par to add some whitespace. 
$^\dagger$ Work done during research internship at Ericsson Research, Bangalore, India.
}

\contribution{
    Provide a succinct but precise list of the contribution(s) of the paper. Use contextual notes to avoid implications of contributions more significant than intended and to clarify and situate the contribution relative to prior work (see the examples below). If there is no additional context, enter ``None''. Try to keep each contribution to a single sentence, although multiple sentences are allowed when necessary. If using complete sentences, include punctuation. If using a single sentence fragment, you may omit the concluding period. A single contribution can be sufficient, and there is no limit on the number of contributions. Submissions will be judged mostly on the contributions claimed on their cover pages and the evidence provided to support them. Major contributions should not be claimed in the main text if they do not appear on the cover page. Overclaiming can lead to a submission being rejected, so it is important to have well-scoped contribution statements on the cover page.
    }
    {
    None
    }

\contribution{
    The submission template for submissions to RLJ/RLC 2026
    }
    {
    Built from previous RLC/RLJ, ICLR, and TMLR submission templates
    }

\contribution{
    \textit{{[}Example of one contribution and corresponding contextual note for the paper ``Policy gradient methods for reinforcement learning with function approximation'' \citep{Sutton2000}.]}\\ This paper presents an expression for the policy gradient when using function approximation to represent the action-value function.
    }
    {
    Prior work established expressions for the policy gradient without function approximation \citep{Williams1992}.
    }

 % Your keywords

\summary{The summary appears on the cover page. Although it can be identical to the abstract, it does not have to be. One might choose to omit the stated contributions in the Summary, given that they will be stated in the box below. The original abstract may also be extended to two paragraphs. The authors should ensure that the contents of the cover page fit entirely on a single page. The cover page does \textbf{not} count towards the 8--12 page limit.

\lipsum[1]
}

\begin{document}

% \makeCover  % Create the cover page
\maketitle  % Make the title section

\begin{abstract}
In cooperative multi-agent reinforcement learning (MARL), from a deployment perspective, it is challenging and expensive to train agents from scratch for each new environment or task.
In this work, we propose \textbf{GCT-MARL}, a transfer learning framework that builds on the multi-view graph contrastive backbone of MAIL \citep{mail}, and augments it with a per-view, adaptively weighted alignment loss and a two-phase training protocol specifically designed for transfer across populations of varying sizes and compositions. We empirically demonstrate that the proposed framework markedly accelerates convergence on the target task relative to from-scratch training, in both homogeneous (within-faction, varying $N$) and heterogeneous (cross-faction and mixed unit-type) transfer scenarios. Furthermore, we show that the framework naturally supports \emph{continual learning} by sequentially chaining the two-phase transfer protocol across a series of related tasks. Overall, this work provides a unified approach to mitigating key limitations in current MARL transfer methods with new insights at both methodological and empirical levels. \textbf{Code: }\url{https://github.com/ainimesh/GCT-MARL}.
\end{abstract}

%=============================================================================
\section{Introduction}
\label{sec:intro}
%=============================================================================
Cooperative multi-agent reinforcement learning (MARL) has recently achieved substantial progress across a variety of real-world application domains, including autonomous driving \citep{xu2024multiagent}, traffic signal control \citep{zhang2024marlens}, fleet management, auction-market mechanisms \citep{auction_market}, and autonomous control \citep{auto_contrl}. Most contemporary MARL algorithms adopt the \emph{centralized training and decentralized execution (CTDE)} paradigm \citep{ctde1, ctde_review} to enhance the quality of decentralized policies. A range of value decomposition methods build on this paradigm \citep{ctde2, ctde_factor2, ctde_fcator, du2025mail}. Despite recent advances, training cooperative MARL agents from scratch continues to suffer from pronounced sample inefficiency. In most settings, agents must independently rediscover effective coordination strategies for each new task, even when those tasks exhibit substantial structural similarity. \emph{Transfer learning} offers a natural approach to mitigating this limitation: a policy acquired on a \emph{source} task should, in principle, facilitate and accelerate learning on a related \emph{target} task. 

In MARL, however, source and target tasks often differ in the number of agents, the dimensionality of local observations and global states, and the per-agent action spaces. Conventional neural network architectures generally assume fixed-dimensional inputs unless explicitly augmented with specialized adaptation mechanisms. As a result, straightforward parameter transfer under such conditions of \emph{population mismatch} is typically ill-posed at the architectural level -- a question studied extensively in single-agent reinforcement learning \citep{transfer_sa, transfer_sa2}, but transfer under population mismatch in cooperative MARL remains relatively underexplored.

Existing MARL transfer methods \emph{lateral transfer with learnable adapters} \citep{lateral_transfer}, \emph{curriculum-based transfer} \citep{curri_transfer} and \emph{attention or transformer-based pooling to handle variable populations}; LA-QTransformer \citep{laq_transfer} require per-task adapter retraining or hand-designed task sequences, and none align source–target representations. This leaving them vulnerable to negative transfer when the source and target coordination dynamics diverge.

Graph-based representation offers a structurally cleaner foundation for this problem. A normalized graph operator with shared per feature weights is permutation-invariant by construction and remains well defined for \emph{any team size} \citep{wu2019sgc}, making the parameter shape independent of the agent population.
This insight has been exploited within single tasks.
Graph Neural Network (GNN) based MARL communication methods \citep{jiang2020graph, niu2021magic, das2019tarmac, li2021dicg} uses message passing to learn coordinated behaviour, and the recent MAIL framework \citep{mail} pushes this further by showing that multi-view graph contrastive learning, with three views over the original graph, feature similarity graph and higher order topological graph, yields high quality message representation within single task. In single-agent RL, graph-based representations have been shown to enable policy transfer across state-action mismatched tasks \citep{turret}. However, to the best of our knowledge, despite offering a naturally compatible solution, GNN based methods are not well explored in transfer for population-varying MARL. 
We argue that the structural properties of the graph-contrastive \citep{attribute_gcl} backbone, MAIL \citep{mail}, which is built on the population invariant backbone SGC \citep{wu2019sgc}, make it a natural fit for the population-mismatch transfer learning scenario. In this work, we attempt to \emph{re-purpose}, multi-view GCL backbone, originally designed for single-task communication as a transferable prior, provided that source-target representations are explicitly aligned during target training. 
Beyond structural invariance, the contrastive objective itself aids transfer: optimizing cosine similarity captures the scale-invariant relational geometry between agents, which stays meaningful across tasks of differing size and reward magnitude.

\textbf{Contributions.} 
We propose \textbf{GCT-MARL}, a framework that equips MAIL \citep{mail} with a per-view, adaptively-weighted alignment loss, and a two phase training protocol designed for population-varying transfer. Our contributions can be summarized as follows:

\begin{itemize}
    \item To the best of our knowledge, GCT-MARL can be seen as a first attempt to establish a graph contrastive transfer framework for MARL.

    \item We introduce a \emph{per-view adaptively-weighted alignment loss} $\mathcal{L}_{xfer}$ for cross-task transfer that learns which contrastive views carries the strongest signal during transfer learning. 

    \item We also show that the proposed framework also naturally extends to \emph{continual learning}, achieving $89.8\%$ final average accuracy and only $-0.125$ average backward transfer on a four-phase SMAC continual sequence.
    
\end{itemize}

% =============================================================================
\section{Related Work}
\label{sec:related}
% =============================================================================
\textbf{Transfer learning in cooperative MARL.} Lateral connection based transfer method MALT \citep{lateral_transfer} assigns multiple pretrained source policies to each target agent through
Gaussian-mixture-based policy assignment and blends their intermediate features via soft-attention-weighted lateral connections; while MALT achieves general cross-task transfer on top of MADDPG \citep{ctde1},
its design assumes an identical layer scheme between source and target policies and provides no explicit representation-level alignment.
Curriculum-based methods like EPC~\citep{curri_transfer} progressively scale up team size through evolutionary or manually designed task sequences, but require intermediate training stages and are vulnerable when the target task deviates from the curriculum trajectory.
Attention and transformer-based methods such as LA-QTransformer \citep{laq_transfer} handle variable
populations by decomposing coordination into level-adaptive coalition patterns via a population-invariant agent network with transformer (PIT); however, its representation is shaped purely by the TD loss, without any contrastive or alignment objective at the feature level.
Other approaches include policy distillation \citep{policy_distil}, and policy-sharing MARL transfer \citep{psmarl}; while complementary, these methods share the same gap. 
Domain adaptation techniques originating in single-agent setting such as DANN \citep{dann}, CORAL \citep{corl} and CycleGAN \citep{cycle_gan}, have been adapted for transfer baseline in MARL but perform poorly as expected because of their design differences. Across all of these methods, the unifying limitations are the absence of a mechanism to align source and target \emph{representations} at the feature level, and none of these uses a graph-contrastive mechanism for transfer. Our work presents a unified direction to address these limitations and offers new insights on both methodological and empirical level. 

\textbf{Graph neural networks in MARL.} GNN-based communication has been widely used to learn structured inter-agent message passing \citep{jiang2020graph, niu2021magic, das2019tarmac, sukhbaatar2016commnet,
sheng2022lsc, liu2020g2anet, li2021dicg}. Most rely on graph attention \citep{velickovic2017gat} or mean aggregation. Recently, MAIL \citep{mail} introduced multi-view graph contrastive learning over three views (original adjacency, kNN feature-similarity, higher-order topology), achieving state-of-the-art results within single tasks. Graph contrastive learning (GCL) more broadly \citep{velickovic2019dgi, peng2020gmi, hassani2020mvgrl, chen2023asp} has emerged  as a self-supervised representation paradigm but has not been studied for cross-task transfer.
Our work extends MAIL's \citep{mail} multi-view GCL to the \emph{cross-task transfer} setting, and identifies which views best survive distributional shift.

\textbf{Continual reinforcement learning.} Continual-RL methods study how to learn a sequence of tasks while preserving prior knowledge \citep{kirkpatrick2017ewc, schwarz2018progress}. Continual MARL is significantly less studied; we contribute a continual-MARL evaluation on SMAC with population-varying tasks and report both forward and backward transfer. 
We emphasize that GCT-MARL is not explicitly designed for continual learning; rather, the continual behaviour emerges naturally from the transfer mechanism, without a dedicated anti-forgetting regularizer (see Sec.~\ref{sec:tranfer_continual}).

%=============================================================================
\section{Preliminaries}
\label{sec:prelim}
%=============================================================================
\textbf{Agent Graph.}
We represent a cooperative multi-agent system at every timestep $t$ as a graph $\mathcal{G}^{(t)} = (\mathcal{V}, \mathcal{E}^{(t)}, \mathbf{X}^{(t)})$ with $N$ agents as nodes $(\mathcal{V} = \{1,\ldots,N\})$, an undirected edge set $\mathcal{E}^{(t)} \subseteq \mathcal{V}\times\mathcal{V}$, and per-agent feature matrix $\mathbf{X}^{(t)} = [x_1^{(t)}, \ldots, x_N^{(t)}]^\top \in \mathbb{R}^{N\times f}$. Edges encodes which agents can exchange messages in $t$ and are induced by environment specific neighborhood relation, typically visibility, communication range or spatial proximity, producing the adjacency $\mathbf{A}^{(t)} \in \{0,1\}^{N\times N}$. The graph is rebuilt at every step; supporting any task by providing its own $\mathbf{A}^{(t)}$, and making the downstream model invariant to the choice.

\textbf{Dec-POMDP and value decomposition (QMIX).}
A cooperative MARL task can be modelled as a Decentralized Partially Observable Markov Decision Process (Dec-POMDP) \citep{oliehoek2012decpomdp}, and can be represented by a tuple $\mathcal{M} = \langle \mathcal{I}, \mathcal{S}, \{\mathcal{A}^i\}, P, R, \{\Omega^i\}, \{O^i\}, N, \gamma \rangle$, where $\mathcal{I} = \{1, \ldots, N\}$ is the set of $N$ agents, $\mathcal{S}$ is the global state space, $\mathcal{A}^i$ is the action space for agent $i$, $P(s'|s,\mathbf{u}): \mathcal{S} \times \mathcal{U} \to \Delta(\mathcal{S})$ is the transition function, $R(s,\mathbf{u}): \mathcal{S} \times \mathcal{U} \to \mathbb{R}$ is the shared reward, $\Omega^i$ is the observation space, $O^i: \mathcal{S} \to \Omega^i$ is the observation function, and $\gamma \in [0,1)$ is the discount factor.
At each timestep $t$, agent $i$ receives local observation $o^i_t = O^i(s_t)$ and selects action $a^i_t$ according to its policy $\pi^i(a^i|\tau^i)$, where $\tau^i = (o^i_1, a^i_1, \ldots, o^i_t)$ is the action-observation history. The objective is to find the joint policy $\pi^*$ maximizing the expected discounted return: $\pi^* = \arg\max_\pi \; \E_{s,\mathbf{u} \sim \pi} \left[ \sum_{t=0}^{\infty} \gamma^t R(s_t, \mathbf{u}_t) \right]$. For credit assignment, we use QMIX \citep{ctde_fcator}: the joint value $Q_{\mathrm{tot}}$ is factorised as a monotonic combination of per-agent values $Q_i(\tau^i, a^i, h^i)$, where $h^i$ is the per-agent GCL message embedding defined in \ref{sec:mail}. Any monotonic value-decomposition mixer would serve in its place.

\subsection{MAIL: Multi-View Graph Contrastive Learning}
\label{sec:mail}
MAIL \citep{mail} is a single-task graph contrastive learning framework for communication in MARL, on top of which we build our transfer mechanism. MAIL augments QMIX \citep{ctde_fcator} with a multi-view graph contrastive learning (GCL) module that learns the inter-agent communication embedding. The pipeline of MAIL is:
\begin{equation}
\underbrace{o_i}_{\text{raw observations of agent $i$}}
\;\xrightarrow{\;\mathrm{MLP} \rightarrow \mathrm{GRU}\;}\;
\underbrace{z_i \in \mathbb{R}^h}_{\text{agent feature}}
\;\xrightarrow{\;\mathrm{GCL}(\mathbf{Z}, \mathbf{A})\;}\;
\underbrace{\mathbf{H}_o[i]}_{\text{message}},
Q_i \;=\; \mathrm{MLP}_q\big([\,z_i \,\Vert\, \mathbf{H}_o[i]\,]\big),
\label{eq:mail-pipe}
\end{equation}
with $\mathbf{Z} = [z_1,\ldots,z_N]^\top \!\in\! \mathbb{R}^{N\times h}$ the matrix of per-agent GRU outputs that feeds the GCL module.
The front-end is a flat MLP whose input width equals the per-task observation dimension $\dim O$; everything from $z_i$ onward operates on fixed width $h$. The GCL module (defined next) reads $(\mathbf{Z}, \mathbf{A})$ and gives the message $\mathbf{H}_o[i]$ for the per-agent Q-head. 
% The Q-head implements MAIL's abstract $Q_i(\tau_i, a_i, h_i)$ by encoding the history $\tau_i$ as the GRU hidden state $z_i$, taking $h_i = \mathbf{H}_o[i]$, and outputting all $|U|$ Q-values jointly as the vector $\mathrm{MLP}_q([z_i \,\Vert\, \mathbf{H}_o[i]])$.

\paragraph{Three views of the agent graph.} MAIL constructs three views over post GRU feature matrix $\mathbf{Z}$ and original adjacency matrix $\mathbf{A}$. \textbf{(1) Original view ($v_o$)}: the environment-induced visibility adjacency $\mathbf{A}$ (degree-normalized to $\mathbf{S}=\widetilde{\mathbf{D}}^{-1/2}(\mathbf{A}+\mathbf{I})\widetilde{\mathbf{D}}^{-1/2}$). This serves as an anchor view. \textbf{(2) Feature view ($v_f$):} a $k$-nearest-neighbor graph $\mathbf{A}_F$ built from cosine distance between rows of $\mathbf{Z}$, connecting agents with similar internal states regardless of physical proximity. Its normalized adjacency is $\mathbf{S}_F = \widetilde{\mathbf{D}}_F^{-1/2}(\mathbf{A}_F + \mathbf{I})\widetilde{\mathbf{D}}_F^{-1/2}$. \textbf{(3) Topological view ($v_t$):} a higher-order graph obtained by raising $\mathbf{S}$ (or $\mathbf{S}_F$) to a power $l > p$, propagating information over a global neighborhood and capturing broader topology.

All three views are encoded with Simple Graph Convolution \citep{wu2019sgc}, which collapses a GCN into a single linear layer applied after $p$ propagation steps: $\mathbf{H} = \mathbf{S}^p\mathbf{Z} \mathbf{W}$.  With three learnable matrices $\mathbf{W}_f, \mathbf{W}_r, \mathbf{W}_t \!\in\! \mathbb{R}^{h\times h}$, MAIL produces four embeddings:
\begin{align}
\mathbf{H}_o &= \mathbf{S}^p\mathbf{Z} \mathbf{W}_f,
&
\mathbf{H}_f &= \mathbf{H}_o + \mathbf{S}_F\mathbf{Z} \mathbf{W}_f,
&
\mathbf{H}_r &= \mathbf{S}^p\mathbf{Z} \mathbf{W}_r,
&
\mathbf{H}_t &= \mathbf{S}^l\mathbf{Z} \mathbf{W}_t,
\quad (l \!>\! p).
\label{eq:mail-views}
\end{align}

Here $\mathbf{W}_f$ is shared between $\mathbf{H}_o$ and $\mathbf{H}_f$ and only the original-view message $\mathbf{H}_o[i]$ is consumed downstream.
$\mathbf{H}_r$ is a parallel re-encoding of $v_o$ with a separate $\mathbf{W}_r$; serving as a second projection used by two of the three losses below. 

\paragraph{GCL objective.} Three views are aligned by InfoNCE \citep{infonce}, a contrastive loss whose positive and negative samples are defined as follows: for any two view embeddings $\mathbf{H}_a, \mathbf{H}_b \!\in\! \mathbb{R}^{N\times h}$, agent $i$'s embeddings in the two views, $h_i^a$ and $h_i^b$, are considered as positive samples, while the embeddings of all other nodes are considered as negative samples. With cosine similarity $D(\cdot,\cdot)$ and temperature $\tau$, the general form of InfoNCE used to contrast different views can be defined as:
\begin{equation}
\mathrm{InfoNCE}(\mathbf{H}_a, \mathbf{H}_b) \;=\;
-\frac{1}{N}\sum_{i=1}^{N}
\log
\frac{
\exp\!\big(D(h_i^a, h_i^b)/\tau\big)
}{
\sum_{j} \exp\!\big(D(h_i^a, h_j^b)/\tau\big) +
\!\!\sum_{v\in\{a,b\}} \sum_{j\neq i} \exp\!\big(D(h_i^v, h_j^v)/\tau\big)
}.
\label{eq:infonce-pre}
\end{equation}
The GCL objective can be defined as:
\begin{equation}
\mathcal{L}_{gcl} \;=\; \mathcal{L}_f \;+\; \lambda_1\,\mathcal{L}_t \;+\; \lambda_2\,\mathcal{L}_c,
\label{eq:mail-gcl}
\end{equation}
where, \emph{feature-preserving} loss $\mathcal{L}_f = \mathrm{InfoNCE}(\mathbf{H}_o, \mathbf{H}_f)$, \emph{topology-preserving} loss $\mathcal{L}_t = \mathrm{InfoNCE}(\mathbf{H}_r, \mathbf{H}_t)$ and \emph{cross-module} consistency loss $\mathcal{L}_c = \mathrm{InfoNCE}(\mathbf{H}_o, \mathbf{H}_r)$. The total single-task training objective combines QMIX's TD loss with the overall graph contrastive loss becomes $\mathcal{L} = \mathcal{L}_{TD} + \beta\,\mathcal{L}_{gcl}$, with $\beta$ trading off TD against the auxiliary contrastive signal.

%=============================================================================
\section{Method: GCT-MARL}
\label{sec:method}
%=============================================================================
GCT-MARL contributes three things on top of the MAIL backbone of Section~\ref{sec:mail}: (1) a per-entity input encoder that removes MAIL's coupling between encoder shape and observation dimension, (2) a per-view, adaptively-weighted cross-task alignment loss $\mathcal{L}_{xfer}$ that aligns target and frozen-source representations, and (3) a two-phase training protocol that chains naturally into continual sequences.

\subsection{Entity Encoder for Population Invariant Input}
\label{sec:entity_encoder}

MAIL applies flat MLP $W \in \mathbb{R}^{d_o \times h}$ to the raw observation $o_i \in \mathbb{R}^{d_o}$, where $d_o$ depends on the per-task entities. A target task with different population yields different $d_o$, making the source backbone structurally incompatible with the target.
In GCT-MARL we propose a simpler, parameter-efficient construction: we replace MAIL's flat MLP with a per-entity encoder ($\mathrm{EntityEnc}_\phi$), whose parameter shapes depend only on per-entity feature widths, never on the number of entities. Once trained on the source, the encoder is reused across any target sharing the same entity-type schema, with no per-task adapter retraining. We assume each observation decomposes into $T$ typed entity blocks, where type $t$ has a fixed per-entity feature width $d_t$ and a variable number of slots $n_t$ (single-slot types reduce to $n_t\!=\!1$). The encoder applies a per-type projection $\phi^{(t)}$ to each slot, masked-mean-pools across the slots of each type, and fuses the per-type pooled embeddings:
\begin{equation}
e_i \;=\;
\mathrm{MLP}_\mathrm{fuse}\!\left(
\,\Big\|_{t=1}^{T}\;
\mathrm{MM}_{j\le n_t}
\big[\phi^{(t)}\!\big(o_{i,j}^{(t)}\big)\big]
\,\right)
\;\in\; \mathbb{R}^{E},
\label{eq:entity-encoder-general}
\end{equation}
where $\phi^{(t)} : \mathbb{R}^{d_t} \to \mathbb{R}^{E}$ is shared across all slots of type $t$, $\|$ denotes concatenation, and the mask is taken from each slot's alive/visible bit. The output $e_i$ has fixed dimension $E$ regardless of $\{n_t\}$, and the per-type mean pool makes the encoder permutation-invariant within each entity type. For single-slot types (e.g.\ self) the pool is a no-op. For clarity of presentation, we use $\mathrm{MM}$ for $\mathrm{MaskedMean}$. We show for instance in SMAC environment \citep{samvelyan2019smac} how we initialize entity encoder. In SMAC each observation has four entity types: self, allies, enemies, and last-action. Per-entity widths $d_\text{self}, d_\text{ally}, d_\text{enemy}, d_\text{move}$ are zero-padded to a configured maximum covering all maps. Eq. \eqref{eq:entity-encoder-general} becomes:

\begin{equation}
e_i \;=\;
\mathrm{MLP}_\mathrm{fuse}\!\left(
\phi_{\mathrm{self}}(o_i^{\mathrm{self}})
\,\Big\|\,
\mathrm{MM}_{j}\big[\phi_{\mathrm{ally}}(o_{i,j}^{\mathrm{ally}})\big]
\,\Big\|\,
\mathrm{MM}_{k}\big[\phi_{\mathrm{enemy}}(o_{i,k}^{\mathrm{enemy}})\big]
\,\Big\|\,
\phi_{\mathrm{move}}(o_i^{\mathrm{move}})
\right),
\label{eq:entity-encoder-smac}
\end{equation}
with $\phi_{\mathrm{self}}, \phi_{\mathrm{ally}}, \phi_{\mathrm{enemy}}$ two-layer MLPs and $\phi_{\mathrm{move}}$ a linear projection (all mapping to $\mathbb{R}^E$). The alive/visible bit shipped with each SMAC ally/enemy row supplies the pool mask; padded slots contribute zero.
% Transferring from $3m$ (each agent sees $2$ ally slots, $3$ enemy slots) to $8m$ ($7$ ally slots, $8$ enemy slots), the same $\phi_{\mathrm{ally}}, \phi_{\mathrm{enemy}}$ weights are reused; the masked mean pool absorbs the slot-count difference and $e_i$ has the same shape on both maps. No per-task module is introduced.
The complete agent pipeline becomes:
\begin{equation}
\begin{aligned}
&\underbrace{o_i}_{\text{raw observations of agent $i$}}
\;\xrightarrow{\mathrm{EntityEnc}_\phi}\;
\underbrace{e_i \in \mathbb{R}^E}_{\text{entity embeddings}}
\;\xrightarrow{\mathrm{GRU}}\;
\underbrace{z_i \in \mathbb{R}^h}_{\text{agent features}}
\;\xrightarrow{\mathrm{GCL}(\mathbf{Z}, \mathbf{A})}\;
\underbrace{\mathbf{H}_o[i]}_{\text{message}}, \\[4pt]
&Q_i \;=\; \mathrm{MLP}_q\big([\,z_i \,\Vert\, \mathbf{H}_o[i]\,]\big).
\end{aligned}
\label{eq:gct_marl_pipeline}
\end{equation}

\subsection{Per-view cross-task alignment $\mathcal{L}_{xfer}$}
\label{sec:cross_loss}
The entity encoder makes the backbone structurally transferable, but does not ensure that target representations carry the same meaning as on the source.
Naive weight transfer (loading backbone trained weights $\theta_S^{\mathrm{bb}}$ into the target and training only on $\mathcal{L}_{TD}^T + \beta\mathcal{L}_{GCL}^T$) allows the loaded model to drift freely under the target task. We close this semantic gap by designing an explicit alignment objective $\mathcal{L}_{xfer}$ that anchors the online target backbone to a frozen source backbone in the representation space.
For every target minibatch we forward the \emph{same} observations through both the target backbone (online weights) and the frozen-source backbone, producing target views $(\mathbf{H}_o^T, \mathbf{H}_f^T, \mathbf{H}_t^T)$ and source views $(\mathbf{H}_o^S, \mathbf{H}_f^S, \mathbf{H}_t^S)$ from (\ref{eq:mail-views}). Our cross-task alignment decomposes into one InfoNCE term per view:
\begin{equation}
\mathcal{L}_{xfer} \;=\;
\alpha_o\,\mathrm{InfoNCE}(\mathbf{H}_o^T,\mathbf{H}_o^S)
\;+\;
\alpha_f\,\mathrm{InfoNCE}(\mathbf{H}_f^T,\mathbf{H}_f^S)
\;+\;
\alpha_t\,\mathrm{InfoNCE}(\mathbf{H}_t^T,\mathbf{H}_t^S),
\label{eq:lxfer}
\end{equation}
with each InfoNCE term defined exactly as in (\ref{eq:infonce-pre}), now pairing the $i$-th target agent's embedding with the $i$-th source agent's embedding.
When $N_T \neq N_S$, pairing uses the first $\min(N_T, N_S)$ rows; since populations are fixed per phase, this clipping is deterministic.
The three weights satisfy the budget constraint $\alpha_o + \alpha_f + \alpha_t = \gamma_{\mathrm{xfer}}$ ($\gamma_{\mathrm{xfer}}\!\ge\!0$, fixed). First-$\min(N)$ pairing is valid up to permutation for homogeneous transfer ($3m\!\to\!8m$), but crude for heterogeneous transfer ($3m\!\to\!1c3s5z$, where only $3$ of nine agents are anchored, to source marines). Since alignment acts on \emph{learned} post-GCL embeddings, the heterogeneous gains show even partial alignment transfers; principled set-based is left for future work.

\paragraph{Adaptive view weighting.} To find out which view carries the most transferable signal we let the model learn the mixing by making the transfer weights $\alpha_o$, $\alpha_f$ and $\alpha_t$ learnable.
\begin{equation}
(\alpha_o,\alpha_f,\alpha_t)
\;=\;
\gamma_{\mathrm{xfer}}\cdot\mathrm{softmax}(\mathbf{a}).
\label{eq:learnable-alpha}
\end{equation}
The softmax keeps $\sum_v \alpha_v = \gamma_{\mathrm{xfer}}$ for any $\mathbf{a}$, so the total cross-task loss magnitude is invariant to the mixing. The logits $\mathbf{a}$ are updated by the same optimizer as the rest of the target network, and are initialized to $\mathbf{0}$, giving a uniform start. The model thus discovers \emph{per task pair} which contrastive view to weight more heavily for transfer.
% — we report which view is favoured empirically in Section \ref{sec:results}.

\paragraph{Target objective.}
The target's total loss combines QMIX's TD error, the within-target contrastive loss inherited from MAIL \citep{mail}, and our cross-task alignment:
\begin{equation}
\mathcal{L}_o^T \;=\;
\mathcal{L}_{TD}^T \;+\; \beta\,\mathcal{L}_{GCL}^T \;+\; \mathcal{L}_{xfer}.
\label{eq:total-target}
\end{equation}

\subsection{Two-phase training and continual learning}
\label{sec:tranfer_continual}
In \textbf{phase 1 — source (single task)}, we train $\theta_S = (\theta_S^{\mathrm{bb}}, \mathrm{MLP}_q^S, \mathrm{Mixer}^S)$ on a source environment $\mathcal{M}_1$ by minimizing MAIL's single-task loss $\mathcal{L}^S = \mathcal{L}_{TD}^S + \beta\,\mathcal{L}_{GCL}^S$. We instantiate the target environment in \textbf{phase 2 — transfer (target task)} with $\theta_S^{\mathrm{bb}}$ and fresh $\mathrm{MLP}_q^T, \mathrm{Mixer}^T$, keep a frozen copy of $\theta_S^{\mathrm{bb}}$ for $\mathcal{L}_{xfer}$, and minimize (\ref{eq:total-target}) on $\mathcal{M}_2$. For the continual learning experiment which emerges naturally from our design choices we move to \textbf{phase 3 — continual learning}. Here, for a sequence $\mathcal{M}_1,\ldots,\mathcal{M}_K$ we apply Phase 2 at every $k\!>\!1$ with $\theta_{k-1}^{\mathrm{bb}}$ — the backbone \emph{from the immediately preceding phase} — as the frozen source.
After every phase $k$ we save the per-phase head $(\mathrm{MLP}_q^k, \mathrm{Mixer}^k)$. To measure \emph{backward} transfer on $\mathcal{M}_j$ ($j\!<\!k$), we pair the saved $\mathrm{MLP}_q^j$ with the current backbone $\theta_k^{\mathrm{bb}}$ and evaluate greedy episodes on $\mathcal{M}_j$. 
This isolates backbone forgetting from action-head forgetting, the latter being unavoidable since each task has its own action space. The off-diagonal entries pair each task's saved head with the \emph{current} backbone, the configuration one would actually deploy, so the reported backward transfer reflects end-to-end retention given a working per-task head, not backbone retention in isolation (see Appendix \ref{app:cl_remark}).

\section{Results and Discussion}
\label{sec:results}
We evaluate on the StarCraft Multi-Agent Challenge \citep{samvelyan2019smac}, using the default sight range and reward scaling. Due to space constrains baselines and evaluation metrics are discussed in Appendix \ref{app:baseline} and experimetal settings in Appendix \ref{app:experimental_setting}. In SMAC we consider multiple map configurations to evaluate the performance while transfer learning including (a) Up-scaling: $3m \to 8m$; (b) Down-scaling: $8m \to 3m$; (c) Heterogeneous transfer: $3m \to 1c3s5z$; (d) Continual learning $3m \to 8m \to 8mvs9m \to 10mvs11m$ .

\begin{figure}[t]
\centering
\begin{subfigure}[t]{0.32\textwidth}
\centering
\includegraphics[width=\linewidth]{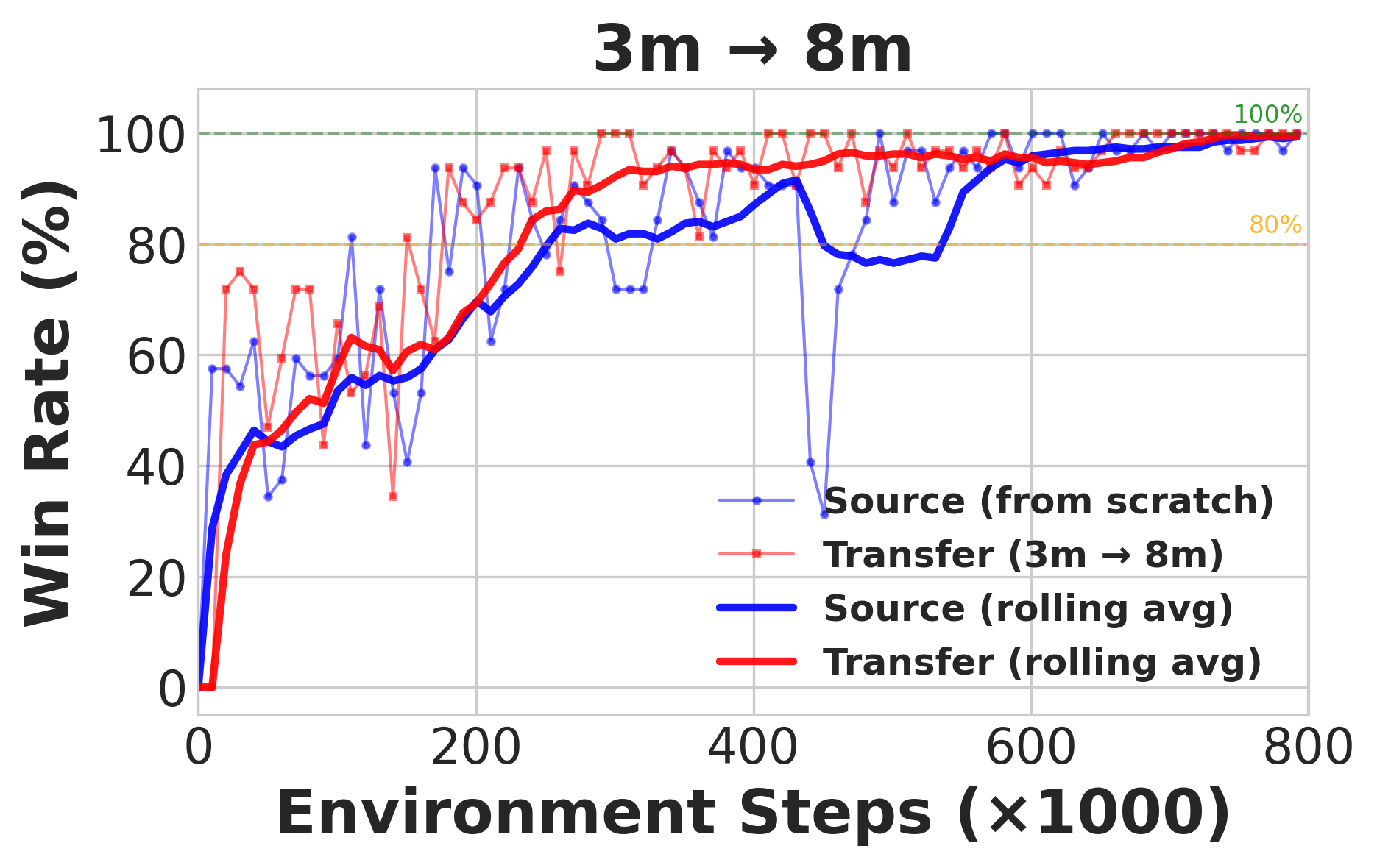}
\caption{3m $\to$ 8m}
\label{fig:transfer-a}
\end{subfigure}
\hfill
\begin{subfigure}[t]{0.32\textwidth}
\centering
\includegraphics[width=\linewidth]{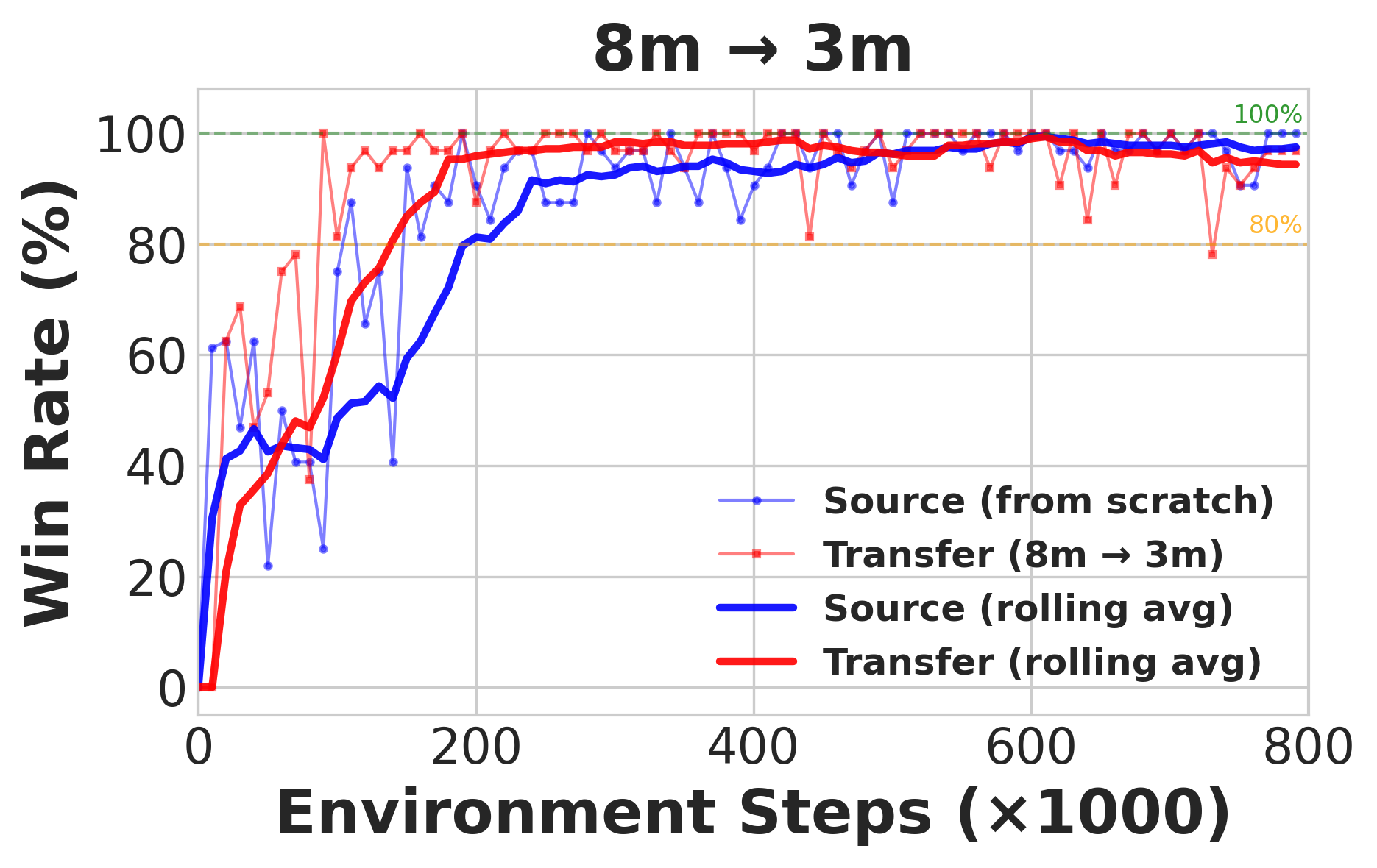}
\caption{8m $\to$ 3m}
\label{fig:transfer-b}
\end{subfigure}
\hfill
\begin{subfigure}[t]{0.32\textwidth}
\centering
\includegraphics[width=\linewidth]{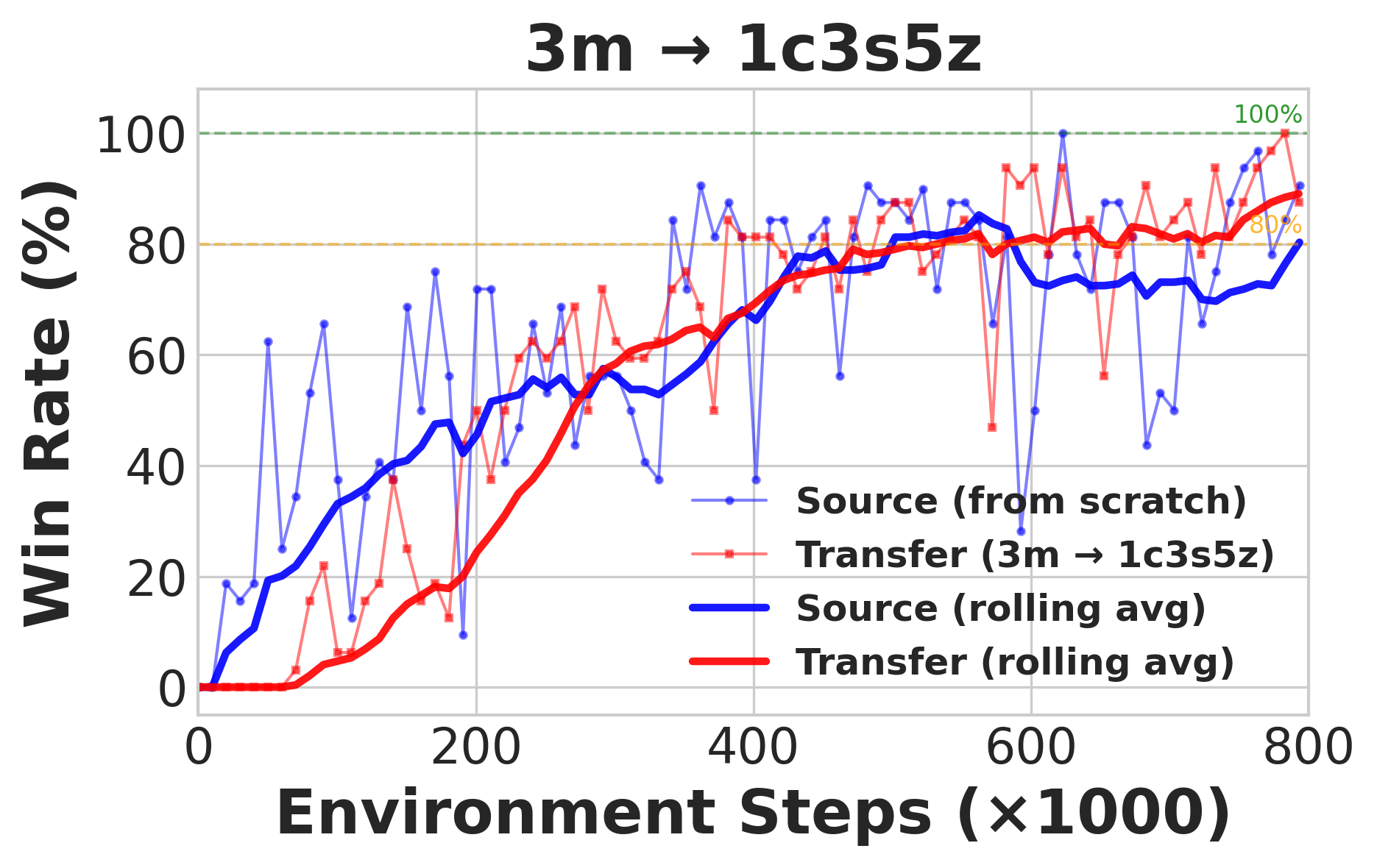}
\caption{3m $\to$ 1c3s5z}
\label{fig:transfer-c}
\end{subfigure}
\caption{Transfer learning performance on SMAC under varying population and composition shifts. Win rate is plotted against environment interaction steps for (a) up-scaling (3m $\to$ 8m), (b) down-scaling (8m $\to$ 3m), and (c) heterogeneous transfer (3m $\to$ 1c3s5z) for best single seed. GCT-MARL consistently achieves faster convergence than baseline methods across all regimes, demonstrating improved sample efficiency under both homogeneous and heterogeneous transfer settings.}
\label{fig:transfer-curves}
\end{figure}

\subsection{Individual Transfers}

Table~\ref{tab:obs-mismatch} reports the four observation-mismatch transfers. 
GCT-MARL reaches the $80\%$ asymptotic reward $2$--$3\times$ faster than from-scratch (MAIL) on the marines maps ($3m\!\to\!8m$, $8m\!\to\!3m$) and on the larger $3m\!\to\!10mvs11m$ and $3m\!\to\!1c3s5z$ transfers, while matching the from-scratch baseline at convergence and report the primary results in this section. The current SOTA method LA-QT \citep{laq_transfer} is still $2$--$3\times$ slower than ours. Per-scenario convergence, using our proposed learnable alpha, curves are shown in Figure~\ref{fig:transfer-curves}.

% =============================================================================
% COMBINED TABLE 1+2: Observation-space mismatch transfer
% =============================================================================
\begin{table}[t]
\caption{Transfer performance under observation-space mismatch in SMAC. Top: number of environment interactions required to reach 80\% of final performance (lower is better). Bottom: final win rate at convergence (higher is better). GCT-MARL significantly reduces sample complexity while maintaining or improving final performance compared to both domain adaptation and MARL transfer baselines.}
\label{tab:obs-mismatch}
\centering
\small
\setlength{\tabcolsep}{0.9pt}
\renewcommand{\arraystretch}{0.90}
\begin{tabular}{l|c|c|c|c|c|c|r}
\toprule
\textbf{Scenario} & \textbf{GCT-MARL} & \textbf{DANN} & \textbf{CORAL} & \textbf{CycleGAN} & \textbf{LA-QT} & \textbf{MAIL} & \textbf{Baseline} \\
\midrule
\multicolumn{8}{c}{\emph{Environment interactions to reach 80\% asymptotic reward ($\times 10^5$)}} \\
\midrule
3m $\to$ 8m              & $\mathbf{1.50 \pm 0.01}$ & $3.1 \pm 0.08$  & $2.8 \pm 0.06$  & $4.1 \pm 0.05$  & $1.72 \pm 0.06$ & $4.91 \pm 0.02$ & $5.1 \pm 0.03$ \\
8m $\to$ 3m              & $\mathbf{0.80 \pm 0.04}$ & $2.8 \pm 0.06$  & $3.3 \pm 0.07$  & $3.1 \pm 0.04$  & $1.42 \pm 0.02$ & $1.90 \pm 0.01$ & $4.82 \pm 0.03$ \\
3m $\to$ 10mvs11m    & $\mathbf{6.40 \pm 0.2}$  & $36.7 \pm 0.5$  & $42.3 \pm 0.3$  & $51.0 \pm 0.4$  & $11.5 \pm 0.4$  & $14.8 \pm 0.3$  & $12.50 \pm 0.1$ \\
3m $\to$ 1c3s5z          & $\mathbf{3.80 \pm 0.03}$ &  $78.7 \pm 0.3$       & $69.4 \pm 0.4$         & $91.5 \pm 0.5$             & $38.7 \pm 0.4$            & $6.22 \pm 0.01$ & $53.4 \pm 0.2$ \\
\midrule
\multicolumn{8}{c}{\emph{Final win rate at convergence}} \\
\midrule
3m $\to$ 8m              & $\mathbf{1.00 \pm 0.00}$ & $0.79 \pm 0.06$ & $0.81 \pm 0.04$ & $0.75 \pm 0.05$ & $0.81 \pm 0.03$ & $1.00 \pm 0.00$ & $0.8 \pm 0.02$ \\
8m $\to$ 3m              & $\mathbf{1.00 \pm 0.00}$ & $0.92 \pm 0.04$ & $0.94 \pm 0.01$ & $0.91 \pm 0.03$ & $0.94 \pm 0.02$ & $1.00 \pm 0.00$ & $0.96 \pm 0.02$ \\
3m $\to$ 10mvs11m    & $\mathbf{0.93 \pm 0.02}$ & $0.65 \pm 0.05$ & $0.68 \pm 0.08$ & $0.66 \pm 0.03$ & $0.75 \pm 0.05$ & $0.93 \pm 0.01$ & $0.77 \pm 0.06$ \\
3m $\to$ 1c3s5z          & $\mathbf{1.00 \pm 0.00}$ & $0.61 \pm 0.06$           & $0.54 \pm 0.04$            & $0.65 \pm 0.03$             & $0.62 \pm 0.04$            & $1.00 \pm 0.00$ & $0.66 \pm 0.03$ \\
\bottomrule
\end{tabular}
\end{table}

\subsection{Continual Learning Scenario}

\begin{figure}[t]
\centering
\includegraphics[width=\linewidth]{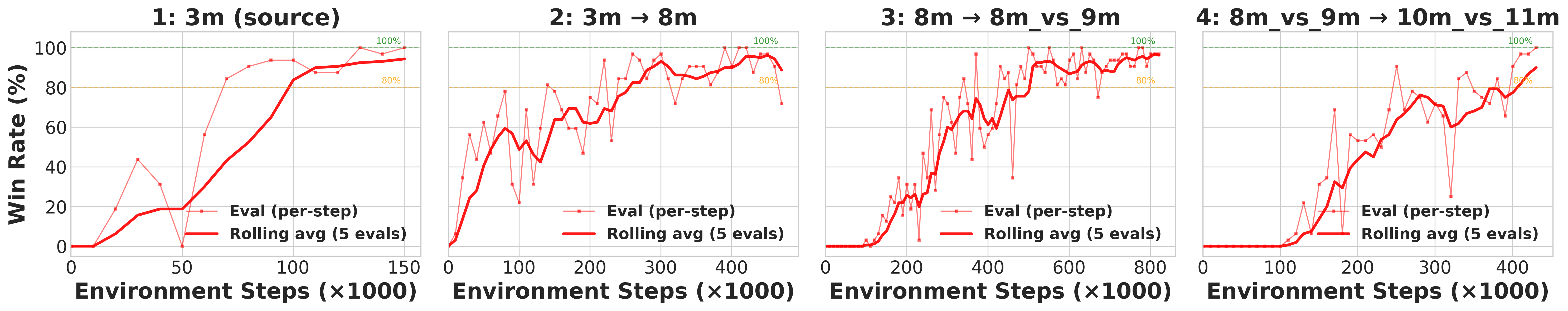}
\caption{Continual learning performance across sequential SMAC tasks. Win rate is shown as a function of environment interactions for the task sequence $3m \to 8m \to 8mvs9m \to 10mvs11m$.}
\label{fig:transfer}
\end{figure}

Table \ref{tab:continual} represents continual evaluation matrix and Figure \ref{fig:transfer} shows the win rate convergence against training steps. In this experiment the backbone for warm-started from previous phase's converged checkpoints, the trained jointly with new Q-head and mixers. The backbone achieves perfect win rate (1.00) on almost every map at the moment it finishes training (diagonal). Forgetting on prior maps is modest: at the end of the full sequence the backbone retains $0.78$ on $3m$, $0.91$ on $8m$, and $0.91$ on $8m{\_}vs{\_}9m$ — an average backward gap of $-0.125$. The most affected task is $3m$, the smallest team, consistent with the entity encoder's larger update pressure when downstream tasks add more entity slots.

\begin{wraptable}{l}{0.5\textwidth}
\vspace{-1.1em}  % tighten against the section heading / prose above
\caption{{\footnotesize Continual learning evaluation across a sequence of SMAC tasks. Each row corresponds to the end of training on a task, with diagonal entries reporting forward performance and off-diagonal entries showing retained performance on previously learned tasks. The model exhibits strong forward transfer and moderate backward transfer degradation, achieving a final average accuracy of 0.898 with limited forgetting.}}
\label{tab:continual}
\centering
\scriptsize
\setlength{\tabcolsep}{3pt}
\renewcommand{\arraystretch}{0.95}
\begin{tabular}{l cccc}
\toprule
End of phase & $3m$ & $8m$ & $\!8m\_vs\_9m\!$ & $\!10m\_vs\_11m\!$ \\
\midrule
1 \;($3m$)               & \textbf{1.00} & ---           & ---           & --- \\
2 \;($8m$)               & 0.91          & \textbf{1.00} & ---           & --- \\
3 \;($8m\_vs\_9m$)   & 0.72          & 1.00          & \textbf{0.97} & --- \\
4 \;($10m\_vs\_11m$) & 0.78          & 0.91          & 0.91          & \textbf{1.00} \\
\midrule
\multicolumn{5}{l}{\textbf{Final ACC} $= 0.898$ \quad
\textbf{Avg BT} $= -0.125$} \\
\bottomrule
\end{tabular}
\vspace{-1em}
\end{wraptable}

\subsection{Ablation}
\begin{figure}[htbp]
\centering
\includegraphics[width=\linewidth]{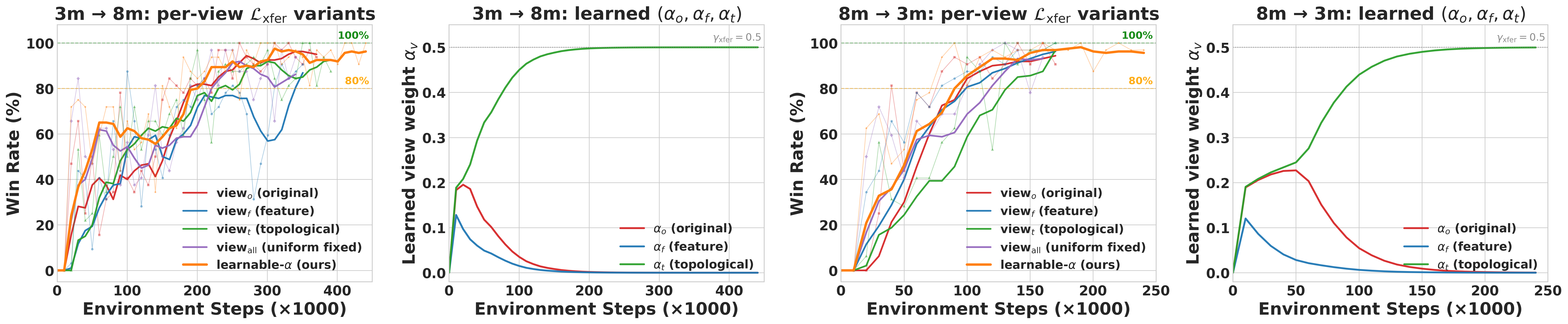}
\caption{Ablation of cross-task alignment strategies. Left: Training curves comparing fixed single-view alignment, uniform multi-view weighting, and the proposed learnable weighting scheme. Right: Evolution of learned alignment weights over training. The adaptive weighting consistently matches or outperforms fixed strategies and converges to favor the topological view.}
\label{fig:ablation}
\end{figure}

Figure \ref{fig:ablation} compares five instantiations of $\mathcal{L}_{xfer}$ on the two marines pairs at seed 0: fixed single-view alignment ($\alpha$ concentrated on $v_o$, $v_f$, or $v_t$ alone), the uniform-fixed mix ($\alpha\!=\!(\gamma_{xfer}/3, \gamma_{xfer}/3, \gamma_{xfer}/3)$), and our learnable-$\alpha$ (eq.~\ref{eq:learnable-alpha}). On both pairs the learnable variant matches or exceeds the best fixed view and dominates the uniform-fixed baseline, indicating that the choice of view-weighting is non-trivial and that committing a priori to any single fixed mix risks underperforming. The bottom row shows that the learned weights converge within $\sim\!50$k env steps to $(\alpha_o,\alpha_f,\alpha_t)\!\approx\!(0,0,\gamma_{xfer})$ on both pairs -- i.e., the \emph{topological} view alone carries essentially all of the transferable signal in this setting -- consistent with prior single-task observations~\citep{mail} that the higher-order topology captures long-range coordination structure that is comparatively task-agnostic.

\section{Conclusions}
We introduced GCT-MARL, the first graph-contrastive transfer framework for cooperative MARL. By recasting multi-view graph contrastive learning as transferable priors with a scale and population-invariant alignment loss and a two-phase protocol, it removes the per-task adapters and population curricula prior work requires, while accelerating target-task learning, mitigating negative transfer, and limiting backward forgetting on SMAC, the last without a dedicated continual-learning component since the transfer objective itself acts as an implicit distillation regularizer. Its main limitation is that the entity encoder assumes a shared entity-type schema between source and target; relaxing this with a schema-agnostic encoder, alongside multi-environment validation, is left for future work.

%%%%%%%%%%%%%%%%%%%%%%%%%%%%%%%%%%%%%%%%%%%%%%%%%%%%%%%%%%%%%%%%
%% Bibliography
%%%%%%%%%%%%%%%%%%%%%%%%%%%%%%%%%%%%%%%%%%%%%%%%%%%%%%%%%%%%%%%%
\newpage
\bibliography{main}

@inproceedings{du2025mail,
  title     = {Multi-Agent Communication with Information Preserving Graph Contrastive Learning},
  author    = {Du, Wei and Ding, Shifei and Guo, Wei and Sun, Yuqing and Yu, Guoxian and Cui, Lizhen},
  booktitle = {Proceedings of the Thirty-Fourth International Joint Conference on Artificial Intelligence (IJCAI)},
  pages     = {64--71},
  year      = {2025},
}

@article{auto_contrl,
  title={Autonomous air traffic controller: A deep multi-agent reinforcement learning approach},
  author={Brittain, Marc and Wei, Peng},
  journal={arXiv preprint arXiv:1905.01303},
  year={2019}
}

@inproceedings{auction_market,
  title={Multi-Agent Reinforcement Learning for Automated Peer-to-Peer Energy Trading in Double-Side Auction Market.},
  author={Qiu, Dawei and Wang, Jianhong and Wang, Junkai and Strbac, Goran},
  booktitle={IJCAI},
  pages={2913--2920},
  year={2021}
}

@inproceedings{wu2019sgc,
  title     = {Simplifying Graph Convolutional Networks},
  author    = {Wu, Felix and Souza, Amauri and Zhang, Tianyi and Fifty, Christopher and Yu, Tao and Weinberger, Kilian},
  booktitle = {Proceedings of the 36th International Conference on Machine Learning (ICML)},
  pages     = {6861--6871},
  year      = {2019},
}

@inproceedings{velickovic2017gat,
  title     = {Graph Attention Networks},
  author    = {Veli{\v{c}}kovi{\'c}, Petar and Cucurull, Guillem and Casanova, Arantxa and Romero, Adriana and Li{\`o}, Pietro and Bengio, Yoshua},
  booktitle = {International Conference on Learning Representations (ICLR)},
  year      = {2018},
}

@inproceedings{velickovic2019dgi,
  title     = {Deep Graph Infomax},
  author    = {Veli{\v{c}}kovi{\'c}, Petar and Fedus, William and Hamilton, William L. and Li{\`o}, Pietro and Bengio, Yoshua and Hjelm, R. Devon},
  booktitle = {International Conference on Learning Representations (ICLR)},
  year      = {2019},
}

@inproceedings{hassani2020mvgrl,
  title     = {Contrastive Multi-View Representation Learning on Graphs},
  author    = {Hassani, Kaveh and Khasahmadi, Amir Hosein},
  booktitle = {Proceedings of the 37th International Conference on Machine Learning (ICML)},
  pages     = {4116--4126},
  year      = {2020},
}

@inproceedings{peng2020gmi,
  title     = {Graph Representation Learning via Graphical Mutual Information Maximization},
  author    = {Peng, Zhen and Huang, Wenbing and Luo, Minnan and Zheng, Qinghua and Rong, Yu and Xu, Tingyang and Huang, Junzhou},
  booktitle = {Proceedings of The Web Conference (WWW)},
  pages     = {259--270},
  year      = {2020},
}

@inproceedings{chen2023asp,
  title     = {Attribute and Structure Preserving Graph Contrastive Learning},
  author    = {Chen, Jialu and Kou, Gang},
  booktitle = {Proceedings of the AAAI Conference on Artificial Intelligence},
  pages     = {7024--7032},
  year      = {2023},
}

@inproceedings{jiang2020graph,
  title     = {Graph Convolutional Reinforcement Learning},
  author    = {Jiang, Jiechuan and Dun, Chen and Huang, Tiejun and Lu, Zongqing},
  booktitle = {International Conference on Learning Representations (ICLR)},
  year      = {2020},
}

@inproceedings{niu2021magic,
  title     = {Multi-Agent Graph-Attention Communication and Teaming},
  author    = {Niu, Yaru and Paleja, Rohan R. and Gombolay, Matthew C.},
  booktitle = {Proceedings of the 20th International Conference on Autonomous Agents and Multiagent Systems (AAMAS)},
  pages     = {964--973},
  year      = {2021},
}

@inproceedings{das2019tarmac,
  title     = {TarMAC: Targeted Multi-Agent Communication},
  author    = {Das, Abhishek and Gervet, Th{\'e}ophile and Romoff, Joshua and Batra, Dhruv and Parikh, Devi and Rabbat, Mike and Pineau, Joelle},
  booktitle = {Proceedings of the 36th International Conference on Machine Learning (ICML)},
  pages     = {1538--1546},
  year      = {2019},
}

@inproceedings{sukhbaatar2016commnet,
  title     = {Learning Multiagent Communication with Backpropagation},
  author    = {Sukhbaatar, Sainbayar and Szlam, Arthur and Fergus, Rob},
  booktitle = {Advances in Neural Information Processing Systems (NeurIPS)},
  pages     = {2244--2252},
  year      = {2016},
}

@inproceedings{sheng2022lsc,
  title     = {Learning Structured Communication for Multi-Agent Reinforcement Learning},
  author    = {Sheng, Junjie and Wang, Xiangfeng and Jin, Bo and Yan, Junchi and Li, Wenhao and Chang, Tsung-Hui and Wang, Jun and Zha, Hongyuan},
  booktitle = {Proceedings of the 21st International Conference on Autonomous Agents and Multiagent Systems (AAMAS)},
  pages     = {436--438},
  year      = {2022},
}

@inproceedings{liu2020g2anet,
  title     = {Multi-Agent Game Abstraction via Graph Attention Neural Network},
  author    = {Liu, Yong and Wang, Weixun and Hu, Yujing and Hao, Jianye and Chen, Xingguo and Gao, Yang},
  booktitle = {Proceedings of the AAAI Conference on Artificial Intelligence},
  pages     = {7211--7218},
  year      = {2020},
}

@inproceedings{li2021dicg,
  title     = {Deep Implicit Coordination Graphs for Multi-Agent Reinforcement Learning},
  author    = {Li, Sheng and Gupta, Jayesh K. and Morales, Peter and Allen, Ross and Kochenderfer, Mykel J.},
  booktitle = {Proceedings of the 20th International Conference on Autonomous Agents and Multiagent Systems (AAMAS)},
  year      = {2021},
}

@ARTICLE{lateral_transfer,
  author={Shi, Haobin and Li, Jingchen and Mao, Jiahui and Hwang, Kao-Shing},
  journal={IEEE Transactions on Cybernetics}, 
  title={Lateral Transfer Learning for Multiagent Reinforcement Learning}, 
  year={2023},
  volume={53},
  number={3},
  pages={1699-1711},
}

@inproceedings{curri_transfer,
  title     = {Evolutionary Population Curriculum for Scaling Multi-Agent Reinforcement Learning},
  author    = {Long, Qian and Zhou, Zihan and Gupta, Abhinav and Fang, Fei and Wu, Yi and Wang, Xiaolong},
  booktitle = {International Conference on Learning Representations (ICLR)},
  year      = {2020},
}

@ARTICLE{laq_transfer,
  author={Zhou, Tianze and Zhang, Fubiao and Shao, Kun and Dai, Zipeng and Li, Kai and Huang, Wenhan and Wang, Weixun and Wang, Bin and Li, Dong and Liu, Wulong and Hao, Jianye},
  journal={IEEE Transactions on Games}, 
  title={Cooperative Multiagent Transfer Learning With Coalition Pattern Decomposition}, 
  year={2024},
  volume={16},
  number={2},
  pages={352-364},
}

@inproceedings{samvelyan2019smac,
  title     = {The {StarCraft} Multi-Agent Challenge},
  author    = {Samvelyan, Mikayel and Rashid, Tabish and de Witt, Christian Schroeder and Farquhar, Gregory and Nardelli, Nantas and Rudner, Tim G. J. and Hung, Chia-Man and Torr, Philip H. S. and Foerster, Jakob and Whiteson, Shimon},
  booktitle = {Proceedings of the 18th International Conference on Autonomous Agents and Multiagent Systems (AAMAS)},
  year      = {2019},
}

@incollection{oliehoek2012decpomdp,
  title     = {Decentralized {POMDP}s},
  author    = {Oliehoek, Frans A.},
  booktitle = {Reinforcement Learning},
  pages     = {471--503},
  year      = {2012},
  publisher = {Springer},
}

@article{kirkpatrick2017ewc,
  title     = {Overcoming Catastrophic Forgetting in Neural Networks},
  author    = {Kirkpatrick, James and Pascanu, Razvan and Rabinowitz, Neil and Veness, Joel and Desjardins, Guillaume and Rusu, Andrei A. and Milan, Kieran and Quan, John and Ramalho, Tiago and Grabska-Barwinska, Agnieszka and others},
  journal   = {Proceedings of the National Academy of Sciences},
  volume    = {114},
  number    = {13},
  pages     = {3521--3526},
  year      = {2017},
}

@inproceedings{schwarz2018progress,
  title     = {Progress \& Compress: A Scalable Framework for Continual Learning},
  author    = {Schwarz, Jonathan and Czarnecki, Wojciech M. and Luketina, Jelena and Grabska-Barwinska, Agnieszka and Teh, Yee Whye and Pascanu, Razvan and Hadsell, Raia},
  booktitle = {Proceedings of the 35th International Conference on Machine Learning (ICML)},
  year      = {2018},
}

@article{xu2024multiagent,
  title     = {A Multi-Agent Reinforcement Learning Based Control Method for Connected and Autonomous Vehicles in a Mixed Platoon},
  author    = {Xu, Yaqi and Shi, Yan and Tong, Xiaolu and others},
  journal   = {IEEE Transactions on Vehicular Technology},
  year      = {2024},
}

@article{zhang2024marlens,
  title     = {{MARLens}: Understanding Multi-Agent Reinforcement Learning for Traffic Signal Control via Visual Analytics},
  author    = {Zhang, Yutian and Zheng, Guohong and Liu, Zhiyuan and others},
  journal   = {IEEE Transactions on Visualization and Computer Graphics},
  year      = {2024},
}

@article{transfer_sa,
  title={Transfer learning for reinforcement learning domains: A survey.},
  author={Taylor, Matthew E and Stone, Peter},
  journal={Journal of Machine Learning Research},
  volume={10},
  number={7},
  year={2009}
}

@article{transfer_sa2,
  title={Transfer learning in deep reinforcement learning: A survey},
  author={Zhu, Zhuangdi and Lin, Kaixiang and Jain, Anil K and Zhou, Jiayu},
  journal={IEEE Transactions on Pattern Analysis and Machine Intelligence},
  volume={45},
  number={11},
  pages={13344--13362},
  year={2023},
  publisher={IEEE}
}

@article{ctde_review,
  title={Deep multiagent reinforcement learning: challenges and directions.},
  author={Wong, Annie and B{\"a}ck, Thomas and Kononova, Anna V and Plaat, Aske},
  journal={Artificial Intelligence Review},
  volume={56},
  number={6},
  year={2023}
}

@article{ctde1,
  title={Multi-agent actor-critic for mixed cooperative-competitive environments},
  author={Lowe, Ryan and Wu, Yi I and Tamar, Aviv and Harb, Jean and Pieter Abbeel, OpenAI and Mordatch, Igor},
  journal={Advances in neural information processing systems},
  volume={30},
  year={2017}
}

@inproceedings{ctde2,
  title={Lazy agents: A new perspective on solving sparse reward problem in multi-agent reinforcement learning},
  author={Liu, Boyin and Pu, Zhiqiang and Pan, Yi and Yi, Jianqiang and Liang, Yanyan and Zhang, Du},
  booktitle={International Conference on Machine Learning},
  pages={21937--21950},
  year={2023},
  organization={PMLR}
}

@article{ctde_fcator,
  title={Monotonic value function factorisation for deep multi-agent reinforcement learning},
  author={Rashid, Tabish and Samvelyan, Mikayel and De Witt, Christian Schroeder and Farquhar, Gregory and Foerster, Jakob and Whiteson, Shimon},
  journal={Journal of Machine Learning Research},
  volume={21},
  number={178},
  pages={1--51},
  year={2020}
}

@inproceedings{ctde_factor2,
  title={Expressive multi-agent communication via identity-aware learning},
  author={Du, Wei and Ding, Shifei and Guo, Lili and Zhang, Jian and Ding, Ling},
  booktitle={Proceedings of the AAAI Conference on Artificial Intelligence},
  volume={38},
  number={16},
  pages={17354--17361},
  year={2024}
}

@inproceedings{mail,
  title={Multi-agent communication with information preserving graph contrastive learning},
  author={Du, Wei and Ding, Shifei and Guo, Wei and Sun, Yuqing and Yu, Guoxian and Cui, Lizhen},
  booktitle={Proceedings of the Thirty-Fourth International Joint Conference on Artificial Intelligence},
  pages={64--71},
  year={2025}
}

@inproceedings{turret,
  title={A transfer approach using graph neural networks in deep reinforcement learning},
  author={Yang, Tianpei and You, Heng and Hao, Jianye and Zheng, Yan and Taylor, Matthew E},
  booktitle={Proceedings of the AAAI conference on artificial intelligence},
  volume={38},
  number={15},
  pages={16352--16360},
  year={2024}
}

@inproceedings{attribute_gcl,
  title={Attribute and structure preserving graph contrastive learning},
  author={Chen, Jialu and Kou, Gang},
  booktitle={Proceedings of the AAAI conference on artificial intelligence},
  volume={37},
  number={6},
  pages={7024--7032},
  year={2023}
}

@article{policy_distil,
  title={Policy distillation},
  author={Rusu, Andrei A and Colmenarejo, Sergio Gomez and Gulcehre, Caglar and Desjardins, Guillaume and Kirkpatrick, James and Pascanu, Razvan and Mnih, Volodymyr and Kavukcuoglu, Koray and Hadsell, Raia},
  journal={arXiv preprint arXiv:1511.06295},
  year={2015}
}

@inproceedings{psmarl,
  title={Zero shot transfer learning for robot soccer},
  author={Schwab, Devin and Zhu, Yifeng and Veloso, Manuela},
  booktitle={Proceedings of the 17th International Conference on Autonomous Agents and MultiAgent Systems},
  pages={2070--2072},
  year={2018}
}

@article{dann,
  title={Domain-adversarial training of neural networks},
  author={Ganin, Yaroslav and Ustinova, Evgeniya and Ajakan, Hana and Germain, Pascal and Larochelle, Hugo and Laviolette, Fran{\c{c}}ois and March, Mario and Lempitsky, Victor},
  journal={Journal of machine learning research},
  volume={17},
  number={59},
  pages={1--35},
  year={2016}
}

@inproceedings{corl,
  title={Return of frustratingly easy domain adaptation},
  author={Sun, Baochen and Feng, Jiashi and Saenko, Kate},
  booktitle={Proceedings of the AAAI conference on artificial intelligence},
  volume={30},
  number={1},
  year={2016}
}

@inproceedings{cycle_gan,
  title={Unpaired image-to-image translation using cycle-consistent adversarial networks},
  author={Zhu, Jun-Yan and Park, Taesung and Isola, Phillip and Efros, Alexei A},
  booktitle={Proceedings of the IEEE international conference on computer vision},
  pages={2223--2232},
  year={2017}
}

@inproceedings{infonce,
  title={Noise-contrastive estimation: A new estimation principle for unnormalized statistical models},
  author={Gutmann, Michael and Hyv{\"a}rinen, Aapo},
  booktitle={Proceedings of the thirteenth international conference on artificial intelligence and statistics},
  pages={297--304},
  year={2010},
  organization={JMLR Workshop and Conference Proceedings}
}

@article{lwf,
  title={Learning without forgetting},
  author={Li, Zhizhong and Hoiem, Derek},
  journal={IEEE transactions on pattern analysis and machine intelligence},
  volume={40},
  number={12},
  pages={2935--2947},
  year={2017},
  publisher={IEEE}
}
\bibliographystyle{rlj}

%%%%%%%%%%%%%%%%%%%%%%%%%%%%%%%%%%%%%%%%%%%%%%%%%%%%%%%%%%%%%%%%
%% Appendices
%%%%%%%%%%%%%%%%%%%%%%%%%%%%%%%%%%%%%%%%%%%%%%%%%%%%%%%%%%%%%%%%
\appendix

\section{Baseline Comparison Methods and Metrics Reported}
\label{app:baseline}

We compared performance across two categories of methods: (i) domain adaptation approaches such as DANN \citep{dann}, CORAL \citep{corl} and CycleGAN \citep{cycle_gan}; and (ii) transfer learning approaches such as MALT \citep{lateral_transfer}, PSMARL \citep{psmarl}, Policy Distillation \citep{policy_distil}, EPC \cite{curri_transfer}, LAQTransformer (LA-QT)\citep{laq_transfer}.
MAIL \citep{mail} is the method upon which we build our method and represents single source training convergence and Baseline presents simple GRU based training in Table \ref{tab:obs-mismatch} and \ref{tab:same-obs-dim}. Concretely, Baseline is standard QMIX with a GRU recurrent agent network, trained from scratch on the target task with no graph-contrastive module, no entity encoder, and no $\mathcal{L}_{xfer}$ -- i.e.\ GCT-MARL with its backbone, entity encoder, and alignment loss removed using the same QMIX hyperparameters as Table \ref{tab:hp}.

\paragraph{Metrics.}
Results reported in Table \ref{tab:obs-mismatch} and Appendix \ref{app:same_dim} Table \ref{tab:same-obs-dim} are compared with those of baseline methods using two evaluation metrics: (i) Environment steps taken to reach $80\%$ of the final converged reward, indicating the learning speed, and (ii) the final value of converged reward. The values reported in the Table \ref{tab:obs-mismatch} and \ref{tab:same-obs-dim} are averaged over five random seeds.

\section{Continual Learning Remark}
\label{app:cl_remark}

\begin{remark}[Relation to distillation-based continual learning]
When the frozen source in $\mathcal{L}_{xfer}$ is the immediately preceding phase's backbone, the alignment term pulls the online representation toward the previous task's representation -- functionally a form of representation-level knowledge distillation, closely related to LwF \citep{lwf} and online-distillation continual methods. We therefore do not claim to mitigate forgetting \emph{without any anti-forgetting mechanism}; rather, no \emph{separate} continual-learning component is required, because the same forward-transfer objective induces a distillation-style regularizer as a byproduct. Since action spaces differ across tasks, each phase stores its own action head; the reported BWT measures retention of the shared backbone under the deployable (saved-head) configuration and excludes no hidden head-forgetting term, as the head is frozen at save time. The trade-off is that head storage grows linearly in the number of tasks: GCT-MARL is continual at the representation level but not constant-memory, and execution requires a task-indexed head.
\end{remark}

\section{Comparison Baseline and Experimental Settings.}
\label{app:experimental_setting}
We compared performance across two categories of methods: (i) domain adaptation approaches and (ii) transfer learning approaches, detailed in Appendix \ref{app:baseline}. All the experiments are conducted, using MAIL default parameters with our $\gamma_{\mathrm{xfer}}=0.5$, on NVIDIA H100 NVL GPU.
We follow MAIL defaults: $k=5$, $p=2$, $l=5$, $\lambda_1=0.2$, $\lambda_2=0.3$, $\beta=0.2$, temperature $\tau=0.5$. 
We set $\gamma_{\mathrm{xfer}}=0.5$. 
Optimization follows PyMARL: RMSProp, learning rate $5\times 10^{-4}$ and a batch of 32 episodes. All experiments run on two NVIDIA H100 NVL GPUs; each SMAC SC2 instance is allocated to one GPU, detailed in Table \ref{tab:hp}. 

\begin{table}[h]
\caption{Hyperparameters used in all experiments.}
\label{tab:hp}
\centering
\small
\begin{tabular}{lc}
\toprule
Symbol & Value \\
\midrule
$h$ (hidden dim) & 64 \\
$E$ (embed dim) & 64 \\
$k$ (kNN) & 5 \\
$p$ (SGC hops) & 2 \\
$l$ (topological hops) & 5 \\
$\lambda_1, \lambda_2$ & $0.2, 0.3$ \\
$\beta$ & $0.2$ \\
$\gamma_{\mathrm{xfer}}$ & $0.5$ \\
$\tau$ (InfoNCE temperature) & $0.5$ \\
Optimiser & RMSProp \\
Learning rate & $5\times 10^{-4}$ \\
Batch (episodes) & 32 \\
Replay buffer (episodes) & 5000 \\
Target update interval & 200 grad steps \\
Discount $\gamma$ & 0.99 \\
$\epsilon$ schedule & $1.0 \to 0.05$ over $50{,}000$ env steps \\
Grad clip & 10.0 \\
Mixer embed dim & 32 \\
\bottomrule
\end{tabular}
\end{table}

\section{Same Observation-Space Dimension Transfer}
\label{app:same_dim}
Table~\ref{tab:same-obs-dim} reports the $8m\!\to\!8m\_vs\_9m$ asymmetric-population transfer, where source and target share observation dimensionality but the target faces one extra enemy unit. GCT-MARL reaches $80\%$ of asymptotic reward $4.2\times$ faster than from-scratch and $4.5\times$ faster than the strongest competitor (MALT), and converges to a perfect $1.00$ win rate while baselines plateau between $0.86$ and $0.91$. Because the entity encoder no-ops when input shapes match, the gain over MALT and PSMARL attributes cleanly to representation-level alignment via $\mathcal{L}_{xfer}$ rather than to parameter-shape compatibility. This isolates the contribution of the cross-task alignment loss from the entity encoder.

% =============================================================================
% COMBINED TABLE: Same-observation-dim transfer
% =============================================================================
\begin{table}[htbp]
\caption{Transfer performance under identical observation-space dimensionality with population mismatch. Despite identical input dimensions, the target task introduces additional agents, requiring coordination generalization. GCT-MARL achieves faster convergence and superior final performance, highlighting the effectiveness of representation-level alignment independent of architectural compatibility.}
\label{tab:same-obs-dim}
\centering
\small
\setlength{\tabcolsep}{1pt}
\renewcommand{\arraystretch}{0.90}
\begin{tabular}{l|c|c|c|c|c|c|r}
\toprule
\textbf{Scenario} & \textbf{GCT-MARL} & \textbf{MALT} & \textbf{PSMARL} & \textbf{Distilled} & \textbf{EPC} & \textbf{MAIL} & \textbf{Baseline} \\
\midrule
\multicolumn{8}{c}{\emph{Environment interactions to reach 80\% asymptotic reward ($\times 10^6$)}} \\
\midrule
8m $\to$ 8mvs9m & $\mathbf{0.53 \pm 0.08}$ & $2.41 \pm 0.05$ & $5.23 \pm 0.07$ & $4.28 \pm 0.03$ & $3.02 \pm 0.04$ & $2.25 \pm 0.02$ & $5.84 \pm 0.02$ \\
\midrule
\multicolumn{8}{c}{\emph{Final win rate at convergence}} \\
\midrule
8m $\to$ 8mvs9m & $\mathbf{1.00 \pm 0.00}$ & $0.89 \pm 0.04$ & $0.91 \pm 0.05$ & $0.86 \pm 0.05$ & $0.89 \pm 0.04$ & $1.00 \pm 0.00$ & $0.94 \pm 0.03$ \\
\bottomrule
\end{tabular}
\end{table}

\end{document}